%% file: main.tex
\documentclass{article}

\usepackage{arxiv}
\renewcommand{\headeright}{}  
\renewcommand{\undertitle}{}  
\usepackage[utf8]{inputenc} 
\usepackage[T1]{fontenc}    
\usepackage[hidelinks]{hyperref}
\usepackage{url}            
\usepackage{booktabs}       
\usepackage{amsfonts}       
\usepackage{nicefrac}       
\usepackage{microtype}      
\usepackage{lipsum}		
\usepackage{graphicx}
\usepackage[numbers,sort&compress]{natbib}
\usepackage{doi}
\usepackage{tabularx}
\usepackage{booktabs}
\usepackage{float}
\usepackage{placeins}
\usepackage{caption}
\title{CRISS: A Retrieval-Augmented AI Chatbot for Assisting Cancer Registrars}

\author{
\href{https://orcid.org/0009-0007-8813-0692}
{\includegraphics[scale=0.05]{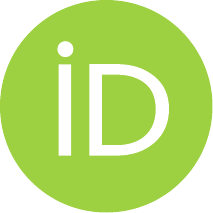}\hspace{0.5mm}\textbf{Vani Seth}}
\textsuperscript{1,a},
\href{https://orcid.org/0000-0002-6095-1743}
{\includegraphics[scale=0.05]{orcid.pdf}\hspace{0.5mm}\textbf{Mohammad Beheshti}}
\textsuperscript{2,3,b},
\href{https://orcid.org/0009-0004-7276-0647}
{\includegraphics[scale=0.05]{orcid.pdf}\hspace{0.5mm}\textbf{Anirudh Kambhampati}}
\textsuperscript{2,3,c},
\\
\href{https://orcid.org/0009-0009-2779-5433}
{\includegraphics[scale=0.05]{orcid.pdf}\hspace{0.5mm}\textbf{Vishwa Bhayani}}
\textsuperscript{2,3,d},
\textbf{Lucinda Ham}
\textsuperscript{2,e},
\href{https://orcid.org/0000-0002-7666-5389}
{\includegraphics[scale=0.05]{orcid.pdf}\hspace{0.5mm}\textbf{Prasad Calyam}}
\textsuperscript{1,f},
\href{https://orcid.org/0000-0002-8159-807X}
{\includegraphics[scale=0.05]{orcid.pdf}\hspace{0.5mm}\textbf{Iris Zachary}}
\textsuperscript{2,3,g,*}
\\[1ex]
\textsuperscript{1}Department of Electrical Engineering and Computer Science,
University of Missouri, Columbia, USA
\\
\textsuperscript{2}Missouri Cancer Registry and Research Center,
Department of Public Health,
\\
College of Health Sciences, University of Missouri, Columbia, USA
\\
\textsuperscript{3}MU Institute for Data Science and Informatics,
University of Missouri, Columbia, USA
\\[1ex]
\textsuperscript{a}vs4ky@missouri.edu \quad
\textsuperscript{b}mbwnh@missouri.edu \quad
\textsuperscript{c}akwg7@missouri.edu
\\
\textsuperscript{d}vbbn32@missouri.edu \quad
\textsuperscript{e}lahf5p@missouri.edu \quad
\textsuperscript{f}calyamp@missouri.edu
\\
\textsuperscript{g}zacharyi@missouri.edu
\\[1ex]
\textsuperscript{*}Corresponding author.
}

\date{}

\hypersetup{
    pdftitle={CRISS: A Retrieval-Augmented AI Chatbot for Assisting Cancer Registrars},
    pdfsubject={Retrieval-Augmented Generation for Cancer Registry Coding Assistance},
    pdfauthor={Vani Seth, Mohammad Beheshti, Anirudh Kambhampati, Vishwa Bhayani, Lucinda Ham, Prasad Calyam, Iris Zachary},
    pdfkeywords={cancer registry, retrieval-augmented generation, large language models, conversational AI, clinical informatics, oncology data specialists, medical question answering, AI safety, clinical decision support}
}

\begin{document}
\maketitle

\begin{abstract}

\textbf{Objectives.} Cancer registrars, including credentialed Oncology Data Specialists (ODSs), must interpret complex and frequently updated cancer registry coding and staging guidelines and standards to abstract accurate data. We developed and evaluated \textit{CRISS (Cancer Registry Intelligent Support System)}, a retrieval-augmented generation (RAG) conversational assistant designed to provide registrars with rapid, source access to relevant guidance. This study evaluated whether CRISS could (1) support accurate, citation-supported responses to registry queries of varying complexity, (2) improve access to and interpretation of relevant guidance, and (3) support training and helpdesk functions while preserving human oversight of final abstraction decisions. 

\textbf{Materials and Methods.} We constructed a domain-specific knowledge base from national standards and cancer registry guidelines, segmented into metadata-tagged passages and indexed as dense vector embeddings. User queries were answered through semantic retrieval of relevant passages followed by citation-grounded response generation with a large language model (LLM), delivered through a React front end and a Flask backend on AWS EC2 with semantic caching. We benchmarked comprising open-weight (e.g., Gemma, Gpt-oss, Qwen), proprietary, and baseline options across the Gemini and GPT families against an easy, medium, and hard registry question set using an LLM-as-a-judge protocol that scored responses on six dimensions (Correctness, Faithfulness, Completeness, Reasoning, Citation Accuracy, Tone) and flagged four failure modes (Hallucination, Clinical Safety, Multi-Part Miss, Rule Conflict). Inference-time retrieval metrics (relevance, faithfulness, hop count, context length, confidence, source year) were logged for every query. 

\textbf{Results.} RAG configurations showed a clear advantage over non-RAG approaches, particularly as question difficulty increased. Mean grounding scores for RAG were 0.62/0.56/0.59 across easy/medium/hard tiers, compared with 0.29/0.26/0.29 for non-RAG, while RAG models also achieved higher semantic-similarity scores overall. In the LLM-as-a-Judge evaluation, proprietary RAG models performed the strongest on easy and medium questions, whereas RAG Local models ranked highest on the hard tier. Follow-up analysis showed that proprietary models were more cautious than local models while answering the questions.

\textbf{Discussion and Conclusion.} The results show that domain-specific RAG substantially improves evidence grounding and response quality for cancer registry questions and supported citation-based responses across varying levels of complexity, with the greatest benefit on more complex cases. Overall, CRISS demonstrates the value of citation-grounded human-centered AI for supporting cancer registrars while preserving human oversight for final coding decisions. 

\end{abstract}


\section{Introduction}
\vspace{-3mm}
Cancer registries are essential to public health surveillance, clinical research, and healthcare planning. By systematically collecting, managing, and analyzing data on cancer incidence, treatment, and survival, registries enable epidemiologic studies, inform policy decisions, guide resource allocation, and support quality improvement initiatives~\cite{cdc_uscs_incidence_sources, tucker2019unlocking}. Hospital-based registries collect detailed information on cancer patients diagnosed and treated within individual institutions to support patient care evaluation and accreditation activities. These cases are subsequently reported to population-based central cancer registries, which aggregate data from multiple facilities within defined geographic areas to produce regional and national statistics that inform surveillance, research, and public health planning~\cite{cdc_npcr_about, seer_overview}. 

The accuracy, timeliness, and completeness of registry data determine the reliability of cancer surveillance~\cite{chen2014review}. These data depend not only on information systems and reporting infrastructure, but also on the expertise of the professionals responsible for interpreting and coding complex clinical information. Oncology Data Specialists (ODSs) abstract diagnostic, staging, treatment, and follow-up information from medical records into standardized registry data formats~\cite{seer_registrars_becoming}. This work extends well beyond data entry and requires advanced interpretive reasoning to reconcile ambiguous documentation, integrate information from multiple sources, and apply complex and evolving coding standards~\cite{weir2020cancer}. Importantly, prior work in cancer registry information management and informatics emphasizes that effective data collection depends on the ability of registrars to interpret complex clinical information, align data abstraction with research and surveillance needs, and navigate evolving data standards, highlighting the cognitive and knowledge-driven nature of registry work~\cite{zachary2015information}.  ODS routinely use national standards and guidelines from organizations such as the North American Association of Central Cancer Registries (NAACCR), the American Joint Committee on Cancer (AJCC), and the Commission on Cancer (CoC)~\cite{amin2017ajcc, edition2004north}. Since relevant guidance is spread across multiple national standard manuals, coding rules, and site-specific guidelines that are periodically revised, resolving a single case often requires consulting multiple references. This knowledge-intensive process can be particularly challenging for newly trained professionals and even trained ODSs as well~\cite{jensen1991cancer}. 

Recent advances in artificial intelligence, particularly Large Language Model (LLM)–based chatbots such as ChatGPT, have demonstrated strong capabilities in natural language understanding, question answering, and knowledge synthesis in healthcare~\cite{beheshti2025evaluating, thirunavukarasu2023large, nazi2024large}. Prior work  has explored trustworthy and secure conversational AI for sensitive healthcare applications, highlighting the importance of reliability, privacy, and safe interaction when LLM-based systems are used to support users in high-stakes domains~\cite{seth2025empathai, seth2026layered, amadi2025building}. These capabilities have motivated substantial research into applying machine learning and NLP to automate clinical information extraction and coding tasks, including the assignment of diagnostic classifications such as the World Health Organization’s International Classification of Diseases (ICD)~\cite{seifaddini2026information}. However, real-world deployment of fully automated systems remains challenging due to variability in medical documentation, the prevalence of rare edge cases, and the potential for even small error rates to produce significant downstream consequences in large datasets~\cite{khalid2026recent,tangudomkit2026artificial}.  

Moreover, LLMs are prone to factual inaccuracies, outdated knowledge, and hallucinated responses, limiting their suitability for unsupervised decision making in critical domains~\cite{huang2025survey}. Retrieval-Augmented Generation (RAG) has emerged as a promising approach to address these reliability concerns by grounding model responses in national standards and guideline source documents retrieved at query time~\cite{lewis2020retrieval}. By combining information retrieval with generative reasoning, RAG systems can reduce hallucinations, enhance factual accuracy, and provide traceable evidence for their outputs~\cite{amugongo2025retrieval}. 

Despite this potential, existing literature reveals a significant gap: current AI-driven approaches overwhelmingly prioritize automation over support, treating registry expertise as a process to be replicated rather than a competency to be augmented~\cite{ji2024unified, chow2026digital, fuchs2026large, dickerson2026fully}. This gap is particularly apparent in retrieval-augmented approaches applied specifically to cancer registry coding, which have similarly centered on automated code generation rather than interactive registrar support and have reported hallucinations and factual inconsistencies even with retrieval grounding in place~\cite{wang2024using}. Across this literature, systems are typically evaluated on output accuracy alone rather than their capacity to enhance user understanding or reinforce guideline comprehension, leaving interactive systems that support registrars' own reasoning largely unexplored. 

In contrast, cancer registry practice requires nuanced interpretation of evolving guidelines, cross-referencing multiple national standards and guidance resources, and careful human judgment across the full abstraction process. To address this need, this work presents a complementary conversational assistant rather than an autonomous coding tool, enabling registrars to ask questions as they arise during abstraction, synthesize rules across documents, and receive explanations supported by explicit citations. By prioritizing transparency, traceability, and human oversight, this paradigm shifts the role of AI from task automation to interactive knowledge support, addressing the practical needs of registry operations where accuracy and accountability are critical. 

This study aims to develop and evaluate \textit{CRISS (Cancer Registry Intelligent Support System)}, a retrieval-augmented conversational assistant designed to support cancer registrars in navigating complex guidelines and resolving registry-related questions. Specifically, the goals are to: (1) determine whether a retrieval-augmented conversational system grounded in national standards and guidelines can provide accurate, citation-supported guidance for real-world queries; (2) assess its potential to improve efficiency in locating and interpreting relevant guidelines; and (3) evaluate its usefulness as a practical support tool for both experienced registrars and those in training while preserving human oversight of final decisions. By reframing AI as an educational adjunct embedded within professional workflows, this study addresses a previously underexplored area of cancer registry informatics: the augmentation of human expertise rather than its replacement.  

The remainder of this paper is organized as follows.  Section~\ref{sec:methods} describes the CRISS system architecture. Section~\ref{sec:eval} presents the evaluation design. Section~\ref{sec:eval-results} discusses the experimental results. Section~\ref{sec:discussion} discusses the findings and limitations. Section~\ref{sec:conclusion} concludes the paper.

\section{Materials and Methods}
\label{sec:methods}
\vspace{-3mm}
This study proposes \textit{CRISS}, a retrieval-augmented conversational system designed to support cancer registrars in navigating, interpreting, and applying complex cancer registry coding guidelines. Registry abstraction requires registrars to interpret clinical documentation, cross-reference multiple national standards and guidance manuals, and apply coding, staging, and reporting standards that are periodically revised and expanded. Rather than functioning as an autonomous coding engine, the system is designed as a human-centered knowledge and educational support tool that assists registrars in locating, interpreting, and applying national standards and  guidance, while final coding decisions remain the responsibility of the human user. 

\begin{figure}[h]
    \centering
    \includegraphics[width=0.9\linewidth]{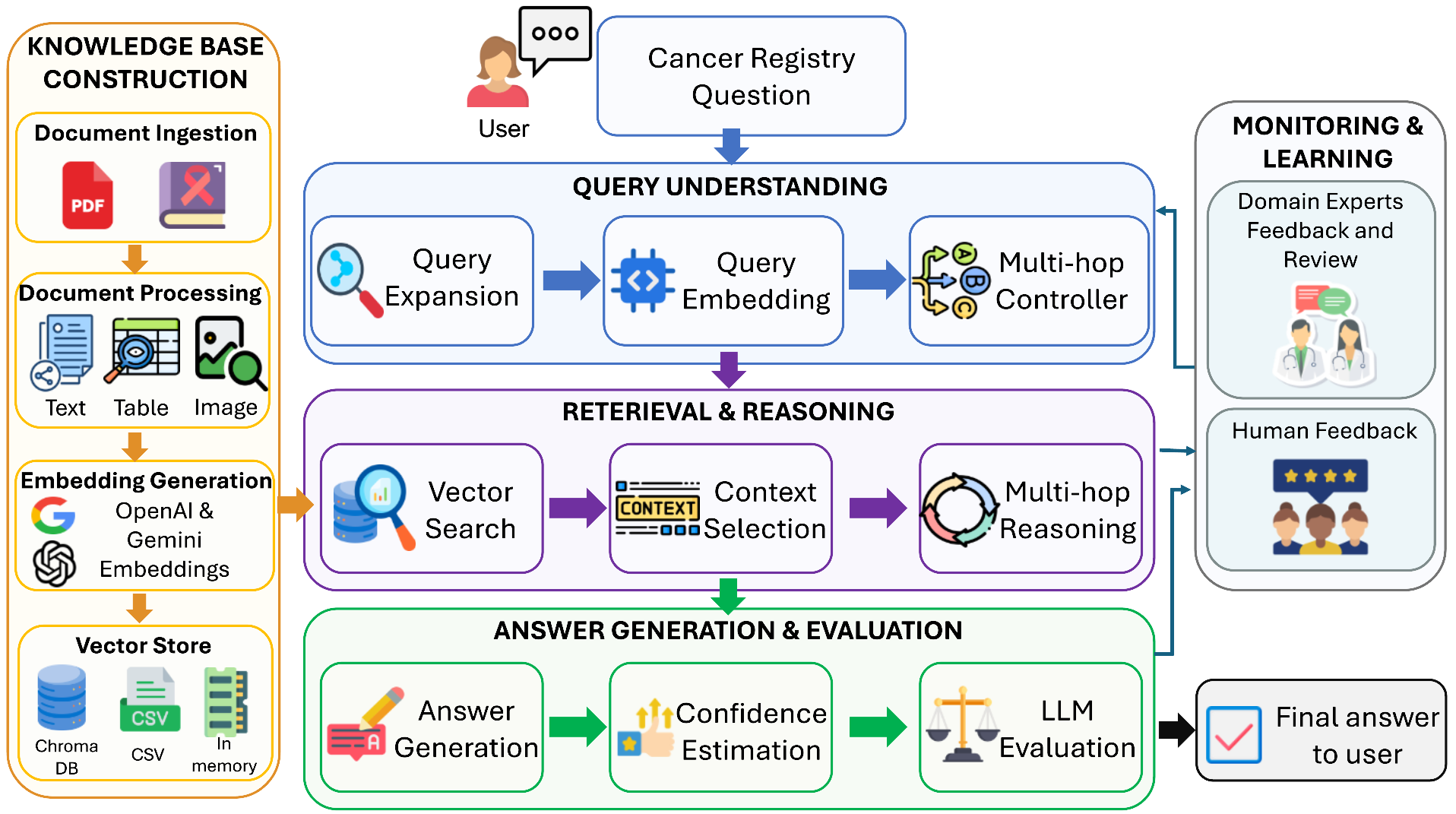}
    \caption{System architecture of CRISS comprising query understanding, multi-hop retrieval and reasoning over a cancer registry knowledge base, and LLM-assisted answer generation with confidence estimation and feedback. }
    \label{fig:system-arch}
\end{figure}

The system adopts a Retrieval-Augmented Generation (RAG) architecture that combines semantic information retrieval with the reasoning capabilities of a Large Language Model (LLM), as illustrated in \textbf{Figure~\ref{fig:system-arch}}. National standards and cancer registry guidelines are converted into a searchable knowledge base. When a registrar submits a natural-language question, the system retrieves passages that are semantically relevant to the query and supplies them to the LLM as supporting context, which then generates a response grounded in that retrieved evidence. This design addresses a key limitation of general-purpose LLM chatbots: their tendency to produce fluent but unsupported or outdated responses. It also allows the knowledge base to be updated independently as coding standards evolve.

The system is guided by three design principles. \textit{First}, every response should be grounded in national standards and cancer registry guideline documentation rather than the model's parametric knowledge, so users can verify the origin of the generated guidance. \textit{Second}, the system prioritizes explanation alongside answer generation, relating retrieved evidence to the user's question rather than producing an isolated recommendation, so it supports both experienced registrars and trainees. \textit{Third}, the system preserves human oversight. It is intended to supplement, rather than replace, registrar expertise by providing evidence, interpretation, and educational support while leaving final coding decisions to the user. 

The remainder of this section describes how these principles are implemented across the system's three components: construction of a registry-specific knowledge base, semantic retrieval of relevant guidelines, and citation-grounded response generation. It then describes the methodology used to evaluate the resulting system.

\subsection{Construction of a Registry-Specific Knowledge Base}
\vspace{-3mm}
The knowledge base was built from national standards and cancer registry guidelines that registrars routinely consult during case abstraction, including cancer coding manuals, Solid Tumor Rules, Multiple Primary and Histology Coding Rules (MP/H) Rules, staging manuals, histology references, and site-specific coding instructions~\cite{seer_coding_manual,seer_solid_tumor_rules,seer_mph_rules,seer_historical_manuals}. Because these documents are distributed across multiple sources and written primarily for human reference, they were preprocessed into a structured representation suitable for semantic search. 

Each source document was parsed to extract its textual content while preserving structural elements such as titles, section headings, rule identifiers, page numbers, tables, and explanatory notes. This context was preserved because many coding decisions depend not only on the wording of an individual rule but also on its surrounding exceptions and document hierarchy; retaining this metadata also allows the system to cite the original source during response generation. 

The extracted text is then segmented into overlapping chunks of 1,500 characters with an overlap of 200 characters, a granularity chosen to preserve coherent rule-and-exception units without exceeding the LLM's context limits. Each resulting chunk is paired with descriptive metadata, including the originating document, section title, page reference, and any available rule identifiers, which is later used to disambiguate conceptually similar passages during retrieval and to support source citation during response generation. Full details of the chunking parameters, the underlying rationale, and the metadata schema are provided in \textbf{Appendix~\ref{app:KB-construction-params}}.

\subsubsection{Semantic Retrieval of Relevant Guidelines}
\vspace{-3mm}
Each text chunk is transformed into a dense semantic representation using the embedding model associated with the selected LLM provider (OpenAI: text-embedding-3-large; Google Gemini: gemini-embedding-001)~\cite{openai_text_embedding_3_large, google_gemini_embeddings}, with OpenAI embeddings used for the open-source model configurations, following prior work demonstrating the effectiveness of dense embedding models for semantic retrieval in medical RAG systems~\cite{klotzman2024enhancing}. The system was designed modularly to support different embedding models and vector-store backends such as csv, chroma, or in-memory, allowing the retrieval configuration to be adapted to the model and deployment environment. Unlike lexical retrieval, semantic embeddings capture contextual meaning, enabling conceptually similar passages to be retrieved even when clinical questions and official coding manuals use different terminology. 

When a registrar submits a question, the same embedding model encodes the query and retrieves the Top-k most relevant passages (k=6 by default; configurable by model). For broad or multi-faceted questions, the system can generate alternative focused search formulations to better capture the individual information needs contained in the original question. Retrieved candidates are additionally reranked to prioritize the most recent applicable registry guidance and de-duplicated before being added to the context. 

Retrieval proceeds through an iterative multi-hop process when a single search is insufficient. After each hop, the system assesses whether the accumulated passages adequately answer the original question. If information is missing, the LLM generates a more focused follow-up query based on the evidence retrieved so far, and another retrieval hop is performed. This process continues until sufficient evidence is obtained or the maximum number of hops is reached. Separating this retrieval layer from the answer generation also allows registry manuals, embedding models, and vector stores to be updated or exchanged without retraining or modifying the overall system. 

\subsection{Citation-Grounded Response Generation}
\vspace{-3mm}
The retrieved passages, together with the user's original question, are combined into a structured prompt supplied to the LLM. The prompt instructs the model to synthesize its response using only the provided registry guidance, explain the reasoning behind its conclusions, and avoid introducing claims beyond the retrieved evidence. Constraining generation to externally retrieved documentation reduces the likelihood of hallucinated responses and keeps generated explanations consistent with national standards and guidelines.  

Response generation is designed to support both experienced registrars and individuals in training, so the objective extends beyond a concise answer. A typical response consists of a direct answer, an explanation of the applicable coding guidance, supporting citations or document references, and contextual information clarifying why the retrieved rule applies to the scenario described. 

Since many registry decisions depend on patient-specific clinical details, the system follows a conservative, evidence-based generation strategy. If the retrieved passages do not provide sufficient information to support a definitive conclusion, the model is instructed to state this limitation explicitly and identify the additional information needed to resolve the question, such as diagnosis date, primary site, histology, tumor behavior, laterality, or prior tumor history, rather than inferring or fabricating an answer. The generated response, together with its supporting citations, is delivered to the registrar through a web-based chatbot interface; interface \& implementation details are provided in \textbf{Appendix~\ref{app:UI}}.

\section{Evaluation}
\label{sec:eval}
\vspace{-3mm}

\subsection{Evaluation Design}
\label{sec:eval-design}
\vspace{-3mm}
\textit{CRISS} was evaluated along two complementary axes: retrieval quality, captured through system-level and retrieval metrics logged at inference time, and response quality, assessed through both automatic metrics and an LLM-as-a-Judge protocol. 

Evaluation was conducted across three difficulty tiers, annotated by experienced subject matter experts (ODSs). Easy questions require direct code lookups, such as ICD-O-3 site and histology codes, with unambiguous ground truth. Medium questions involve multi-step reasoning across coding manuals such as the STORE Manual and Solid Tumor Rules. Hard questions reflect real-world case scenarios submitted by cancer registrars, often involving ambiguous terminology, conflicting rules, or incomplete clinical information. 

To contextualize the contribution of the retrieval pipeline, model performance was additionally benchmarked against a non-RAG baseline condition. In this condition, frontier models from the GPT and Gemini families were queried directly through their standard APIs, without access to the system's retrieval pipeline or curated registry knowledge base. To avoid unfairly handicapping these baseline models, web browsing was enabled through each model's API, allowing them to search the internet for relevant guidance in place of the curated knowledge base. This comparison isolates the performance contribution of the system's domain-specific retrieval pipeline relative to a general-purpose frontier model with generic web-search access.

\subsection{Retrieval Evaluation}
\label{sec:ret-eval}
\vspace{-3mm}
The retrieval pipeline logged six metrics at inference time to capture system-level behavior: confidence score, reflecting the model's self-reported certainty; relevance rating, measuring how well the retrieved context matches the query; faithfulness rating, assessing whether the answer is grounded in the retrieved sources; hop count, indicating the number of retrieval steps performed; context length, measuring the total tokens of context consumed; and reasoning trace, capturing the intermediate steps taken before producing a final answer. Source utilization was also recorded via sources used and a corresponding summary, along with the year of the cited guidelines, to assess the temporal relevance of the referenced rules. 

\subsection{Response Evaluation}
\label{sec:response-eval}
\vspace{-3mm}
\textbf{Automatic Metrics:} Response quality was assessed using standard automatic Natural Language Generation (NLG) metrics such as ROUGE-1/2/L~\cite{lin2004rouge}, BERTScore F1~\cite{zhang2019bertscore}, and sentence-level cosine similarity~\cite{gomaa2013survey}. Full metric definitions, rationale, and results are reported in \textbf{Appendix~\ref{app:rubric-results}}. 

\textbf{LLM-as-a-Judge Protocol:} Beyond automatic metrics, response quality was further evaluated using an LLM-as-a-Judge protocol, following prior work establishing the validity of LLM-based assessment for open-ended generation tasks~\cite{zhou2025automating, croxford2025evaluating}. This approach was adopted over human expert annotation for three reasons. \textit{First}, the volume of evaluations required across 22 generator models,   comprising open-weight/local (e.g., Gemma~\cite{team2024gemma}, gpt-oss~\cite{agarwal2025gpt}, Qwen~\cite{yang2025qwen3}), proprietary, and baseline options across the Gemini and GPT families~\cite{team2023gemini, afridi2026gpt} for three difficulty tiers, and two judge families makes expert annotation prohibitively expensive and time-consuming. \textit{Second}, prior work has demonstrated that capable LLM judges achieve high agreement with domain expert raters on structured evaluation rubrics, particularly when the rubric dimensions are clearly defined and the judge is provided with ground truth for reference. \textit{Third}, LLM judges offer greater consistency than human annotators on repeated evaluations of the same response, reducing noise in the quality signal. 

To reduce the well-documented tendency of LLM judges to rate outputs from their own model family more favorably, a cross-family judging protocol was used. All model outputs were independently evaluated by both GPT-4o and Gemini 3.1 Pro Preview \cite{hurst2024gpt, google_gemini_3_1_pro_preview}, enabling assessment of inter-judge agreement across models and difficulty tiers. Final scores were reported as the mean across both judges, an ensembling approach shown to produce more reliable aggregate scores than single-judge evaluation by reducing the influence of individual judge mannerisms. Models for which the two judges disagreed substantially, defined as a composite score difference exceeding 0.5 points, were flagged for qualitative review, since large disagreements signal that response quality is sufficiently ambiguous to warrant closer inspection. 

The judge evaluated each response across six dimensions scored on a 1 to 5 scale: Correctness (factual accuracy relative to ground truth), Faithfulness (whether the response is grounded in real coding rules rather than hallucinated content), Completeness (whether all components of the question are addressed, particularly for compound questions), Reasoning (the logical soundness of the explanation, independent of whether the final answer is correct), Citation Accuracy (whether cited pages and documents actually support the stated conclusion), and Tone (appropriateness for a teaching and helpdesk context). Extended definitions and the rationale for each dimension are provided in \textbf{Appendix~\ref{app:rubric-results}}. 

\section{Results}
\label{sec:eval-results}
We evaluated \textit{CRISS} following the evaluation design described in the Evaluation (Section~\ref{sec:eval}), analyzing response quality across question difficulty tiers and model configurations.

\subsection{Retrieval Quality}
\begin{figure}[h]
    \centering
    \includegraphics[width=0.7\linewidth]{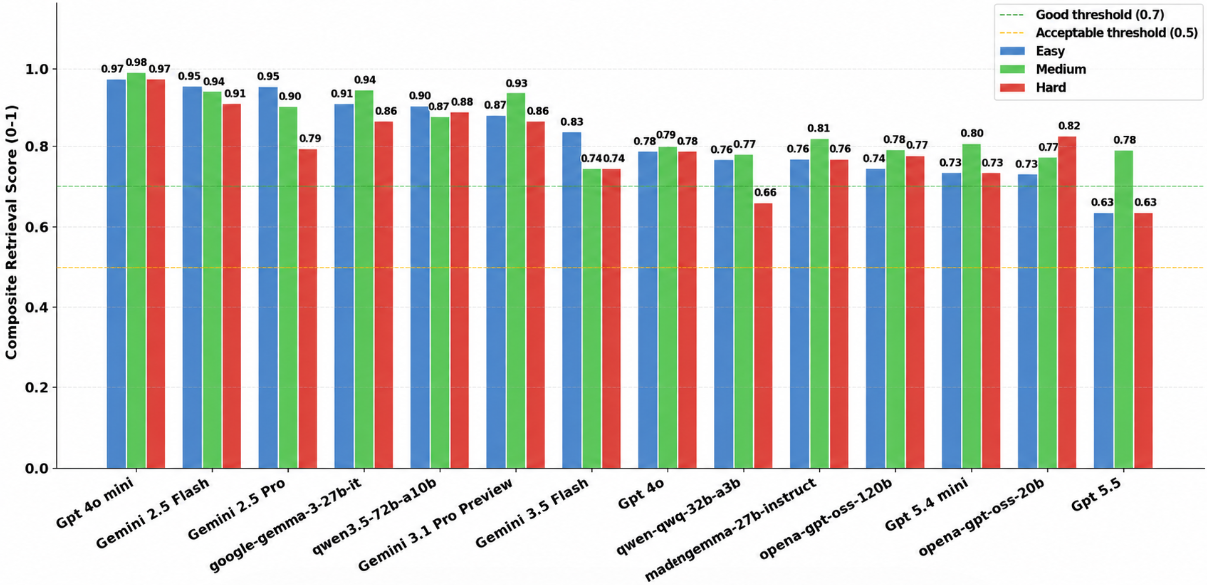}
    \caption{Composite Retrieval Score per model and tier (Sorted in ascending order based on easy tier score).}
    \label{fig:ret-scores}
\end{figure}

Retrieval quality remained high for most RAG configurations across the three difficulty tiers. To summarize retrieval performance, we report a Composite Retrieval Score as shown in \textbf{Figure~\ref{fig:ret-scores}}, which provides a single measure of overall retrieval quality by jointly capturing whether the system retrieved relevant content, remained faithful to the retrieved evidence, and maintained confidence in the retrieved context. The score is calculated as the mean of normalized relevance, faithfulness, and confidence. Composite retrieval scores exceeded the 0.70 reference threshold for nearly all evaluated models across the three tiers. The highest and most consistent scores were observed for GPT-4o-mini (0.97/0.98/0.97 for easy/medium/hard), followed by Gemini-2.5-Flash (0.95/0.94/0.91) and Gemini-2.5-Pro (0.95/0.90/0.79). Google-Gemma-4-31B-QAT (0.91/0.94/0.86) and Qwen3.5-122B-A10B (0.90/0.87/0.88) also maintained strong performance across tiers. Among the remaining models, most scores remained above the 0.70 threshold, although Qwen3.6-35B-A3B decreased to 0.66 on the hard tier, while GPT-5.5 recorded the lowest easy and hard scores at 0.63, despite improving to 0.78 on the medium tier. Overall, the results indicate that retrieval quality was generally preserved as question difficulty increased, although the degree of stability varied across models. 

\subsection{Response Quality}
\begin{figure}[h]
    \centering
    \includegraphics[width=0.8\linewidth]{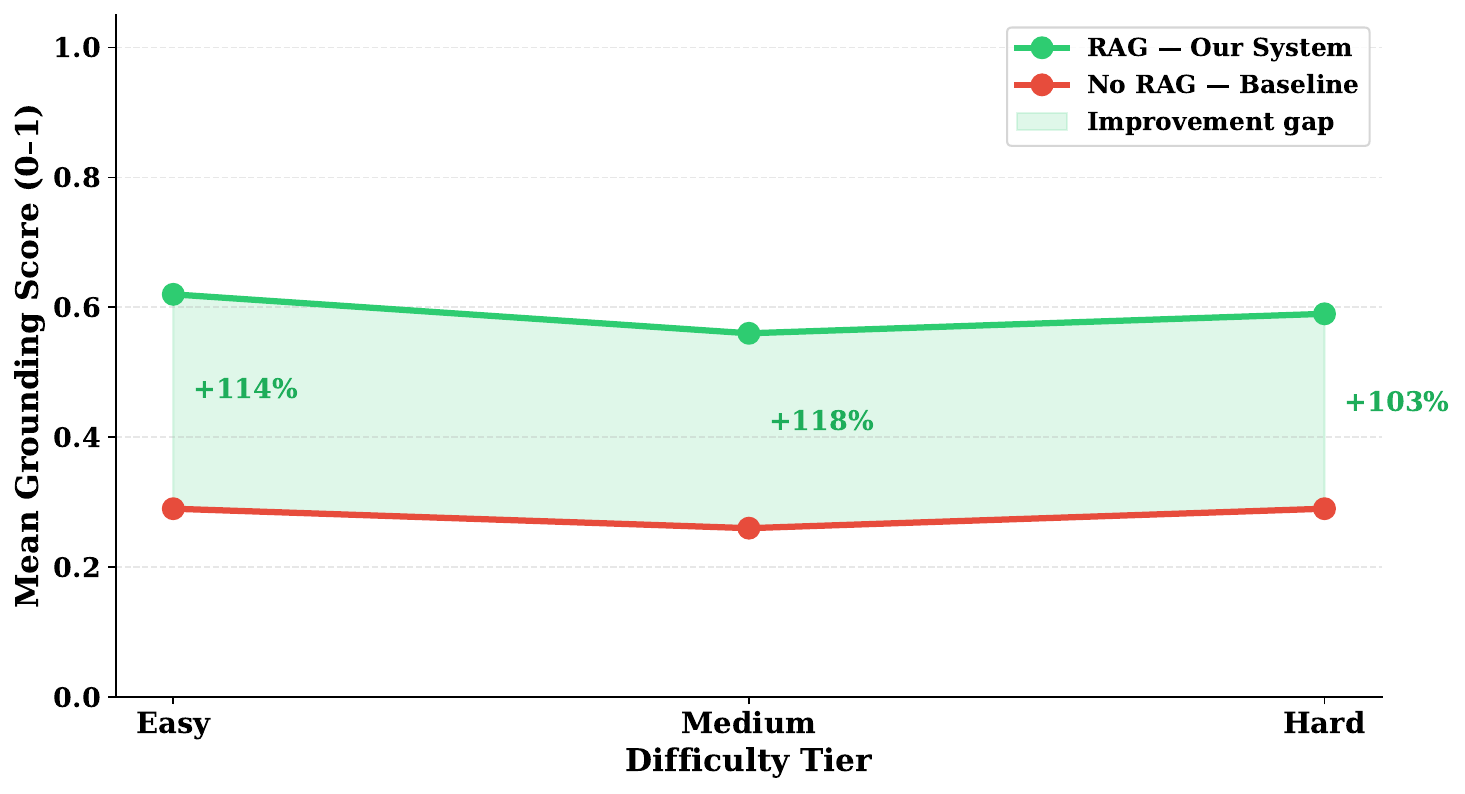}
    \caption{RAG vs. No-RAG Grounding Score Across Difficulty Tiers }
    \label{fig:rag-nonrag}
\end{figure}

To check the need for a RAG-based system, we compared the grounding performance of the RAG configuration with a No-RAG baseline across all three difficulty tiers as shown in \textbf{Figure~\ref{fig:rag-nonrag}}. We report a grounding score, which measures how well the claims in a generated answer are supported by the evidence retrieved by the model. For the RAG system, this evidence comes from the curated registry passages, while the No-RAG baseline uses web-search results. The RAG system achieved mean grounding scores of approximately 0.62, 0.56, and 0.59 for easy, medium, and hard questions, compared with 0.29, 0.26, and 0.29 for the No-RAG baseline. This represents improvements of 114\%, 118\%, and 103\% across the three tiers. These results show that RAG makes the generated answers much more evidence-supported and traceable. The consistent improvement across all difficulty levels supports the need for a RAG-based system for cancer registry coding. 

\begin{figure}[h]
    \centering
    \includegraphics[width=0.8\linewidth]{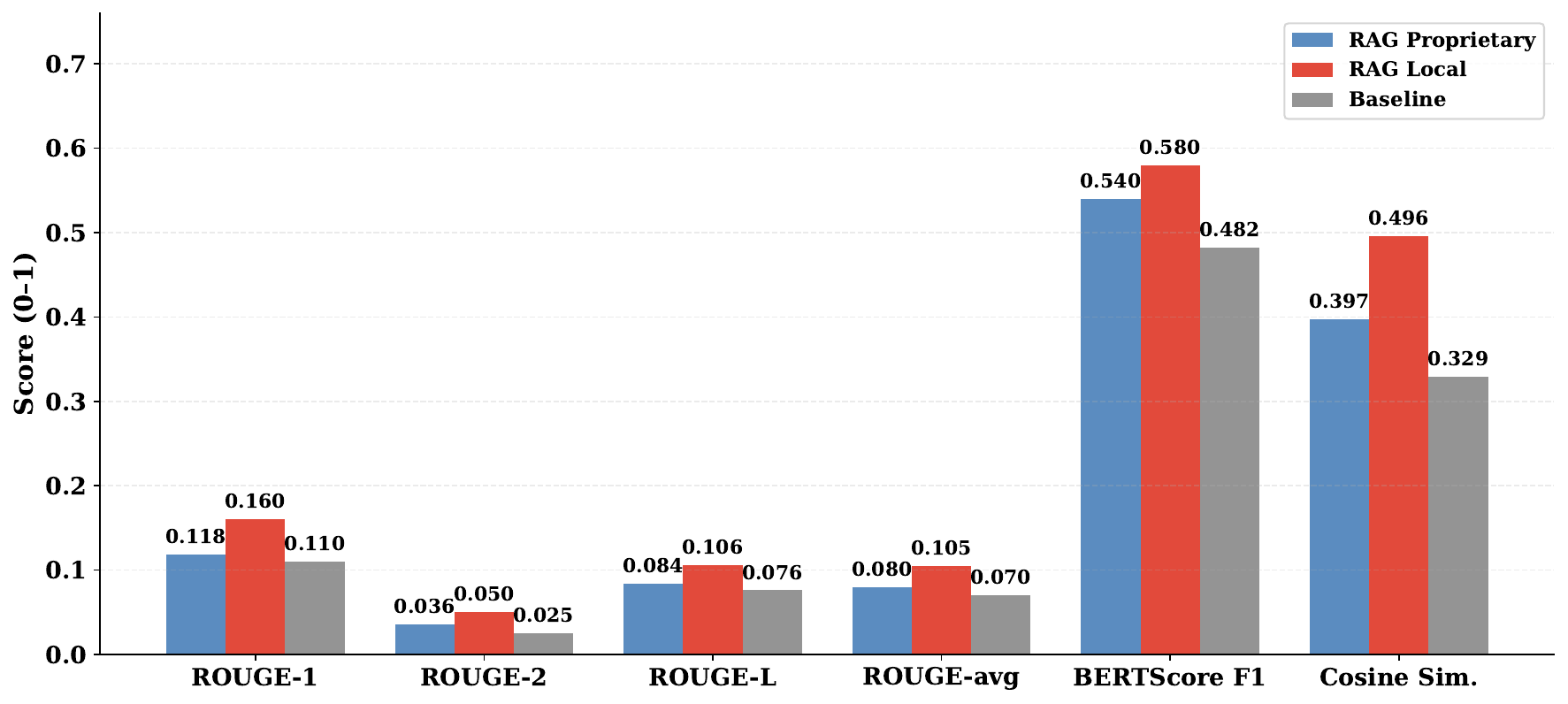}
    \caption{Text Similarity Metrics by Model Groups (Mean across all tiers)}
    \label{fig:text-similarity}
\end{figure}

To compare the overall similarity of generated answers with the reference answers, we report ROUGE, BERTScore F1, and cosine similarity averaged across all difficulty tiers as shown in \textbf{Figure~\ref{fig:text-similarity}}. Across all metrics, the RAG Local models achieved the highest average scores, followed by the RAG Proprietary models, while the baseline models consistently showed the lowest performance. RAG Local achieved a ROUGE average of 0.105, compared with 0.080 for RAG Proprietary and 0.070 for the baseline. A similar pattern was observed for semantic similarity: RAG Local achieved the highest BERTScore F1 of 0.580 and cosine similarity of 0.496, compared with 0.540 and 0.397 for RAG Proprietary and 0.482 and 0.329 for the baseline, respectively. The higher BERTScore and cosine values relative to the ROUGE scores also indicate that the generated answers often captured the same meaning as the reference answers without necessarily using the exact same wording. Overall, the results show that the RAG configurations produced responses that were more closely aligned with the reference answers than the baseline, with the RAG Local models showing the strongest overall similarity. Full ROUGE-1/2/L, BERTScore F1, sentence-level cosine similarity, and associated automatic metric results are reported in \textbf{Appendix~\ref{app:automatic-mertics-results}}. 

\subsection{LLM-as-a-Judge Assessment }
\begin{figure}[h]
    \centering
    \includegraphics[width=\linewidth]{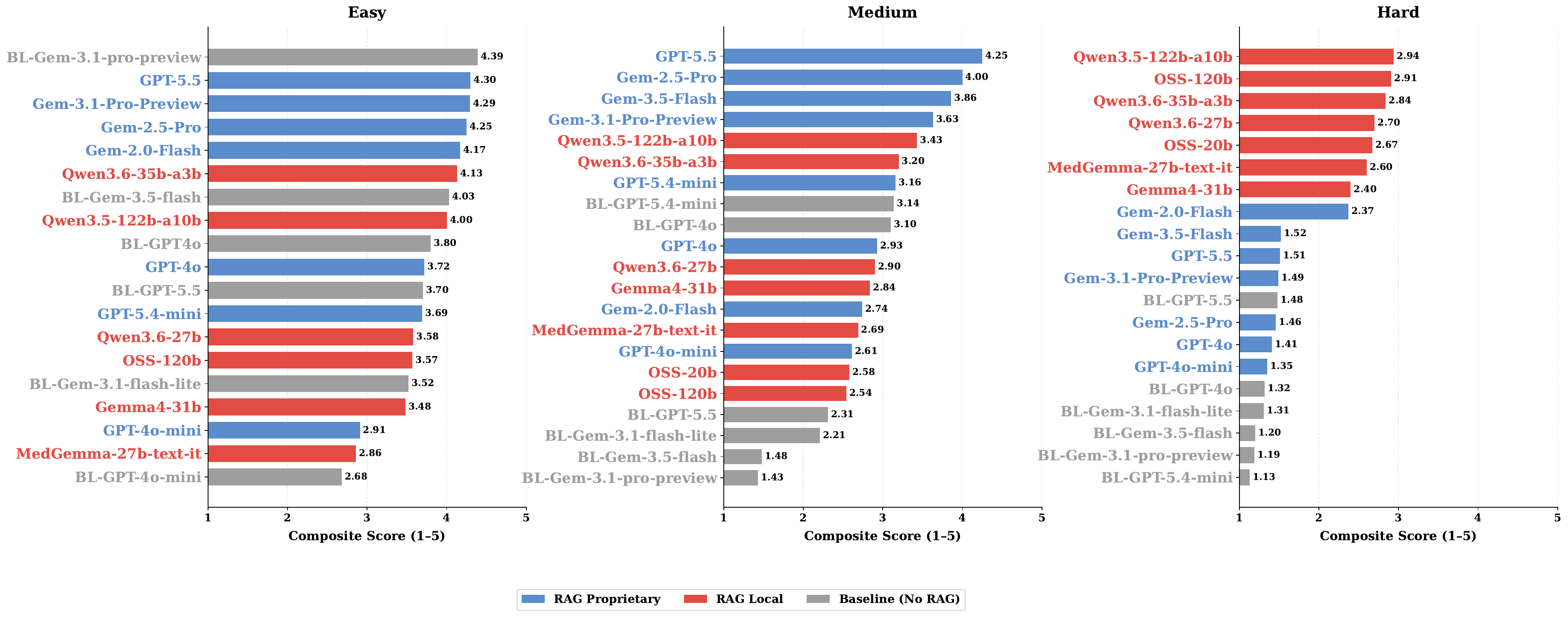}
    \caption{LLM-as-a-Judge Scores for all difficulty tiers}
    \label{fig:llm-judge-tiers}
\end{figure}

To evaluate response quality across all questions and models, we employed an LLM-as-a-Judge method and report an ensemble composite score from GPT and Gemini judges on a 1–5 scale as shown in \textbf{Figure~\ref{fig:llm-judge-tiers}}. Scores from the two judges were averaged at the response level. The complete evaluation rubric used by the judges is provided in \textbf{Appendix~\ref{app:llm-judge-rubric}}.  Performance was strongest on the easy tier, where several models achieved scores above 4.0. The highest score was obtained by Baseline-Gemini-3.1-Pro-Preview (4.39), followed closely by the RAG proprietary models GPT-5.5 (4.30), Gemini-3.1-Pro-Preview (4.29), and Gemini-2.5-Pro (4.25). On the medium tier, GPT-5.5 (4.25) achieved the highest score, followed by Gemini-2.5-Pro (4.00) and Gemini-3.5-Flash (3.86). Scores decreased substantially on the hard tier for most models. The RAG Local models however occupied the highest positions, led by Qwen3.5-122B-A10B (2.94), OSS-120B (2.91), and Qwen3.6-35B-A3B (2.84).  

\section{Discussion}
\label{sec:discussion}

\subsection{Principal Findings}
The results show a clear advantage of RAG over non-RAG approaches, particularly as question difficulty increases. While some non-RAG models remained competitive on easier questions, their performance declined more sharply on medium and hard cases, whereas the RAG configurations generally maintained stronger grounding, semantic similarity, and LLM-as-a-Judge scores. This supports the need for retrieval when answering cancer registry questions that depend on specific coding rules and current reference materials. 

\subsection{Ablation Study: Model Behavior and Response Caution}
\begin{figure}[h]
    \centering
    \includegraphics[width=\linewidth]{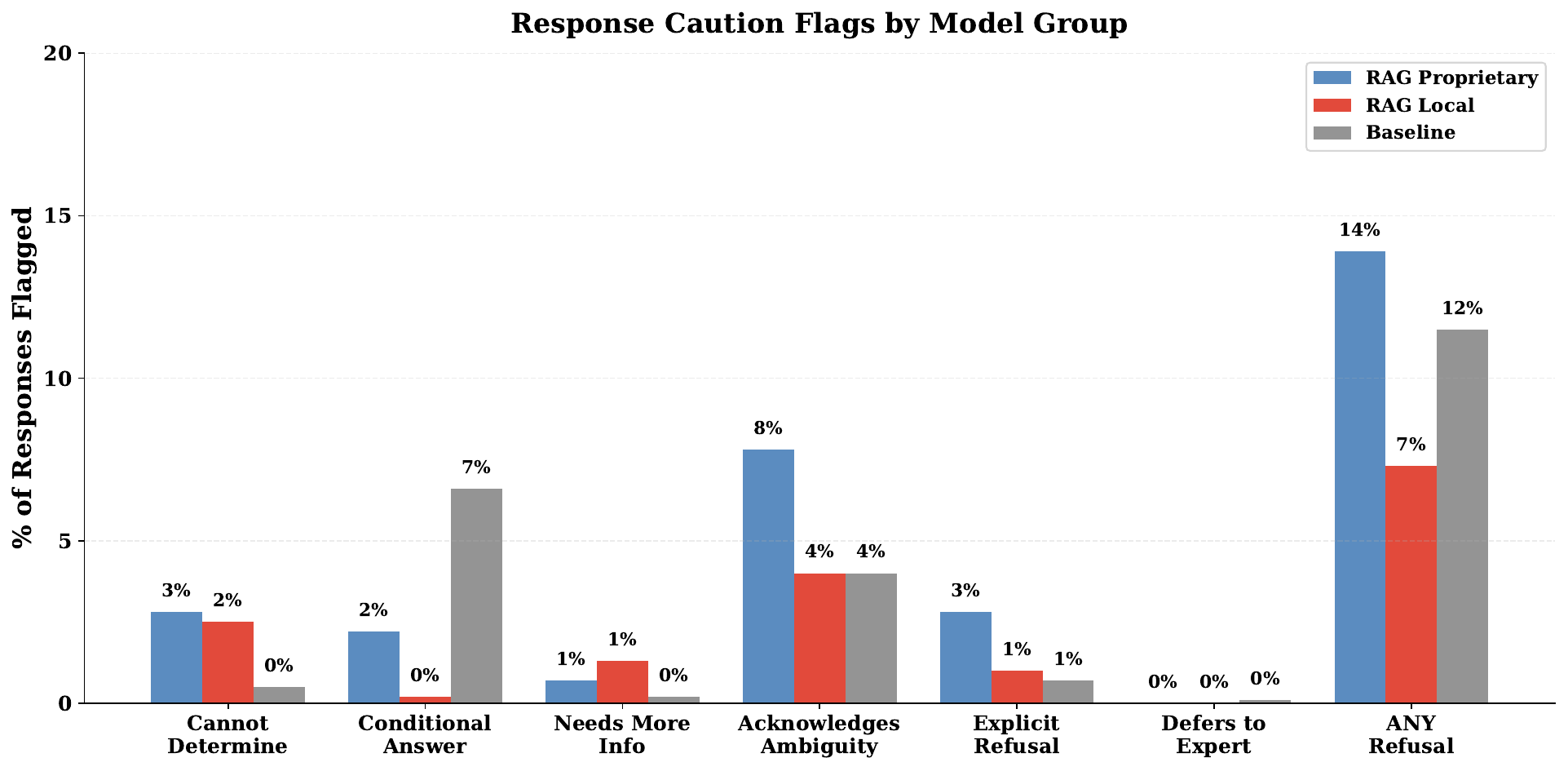}
    \caption{Ablation Study of Response Caution Flags by Model Types}
    \label{fig:ablation-study}
\end{figure}

The LLM-as-a-Judge results revealed a clear shift in model performance as question difficulty increased. While proprietary RAG models performed strongly on the easy and medium tiers, the RAG Local models ranked ahead of the proprietary models on the hard tier. To understand what may be driving this difference, we analyzed model response behavior, including how often models acknowledged ambiguity, requested more information, or refused to provide a definitive answer. The full caution-flag definitions and detailed analysis are provided in\textbf{ Appendix C4}. As shown in \textbf{Figure~\ref{fig:ablation-study}}, RAG Proprietary models showed a higher overall caution/refusal rate than RAG Local models (23.1\% vs. 15.3\%), while local models achieved higher containment (0.401 vs. 0.288) and stronger text-similarity scores, including BERTScore F1 (0.580 vs. 0.540) and cosine similarity (0.496 vs. 0.397). as discussed in \textbf{Figure~\ref{fig:text-similarity}}. These results suggest that proprietary models are more likely to respond cautiously to difficult or ambiguous cases by acknowledging uncertainty, qualifying their response, or avoiding a definitive coding decision. In contrast, local models were more likely to provide a direct coding answer, which increased their overlap with the reference answers and likely contributed to their higher hard-tier scores. Importantly, this does not necessarily mean that the local models are clinically better. Their stronger ranking may partly reflect that they are less likely to acknowledge uncertainty or ambiguity, and therefore more likely to commit to an answer even when the case is difficult. Thus, the hard-tier advantage of local models should be interpreted as a difference in response behavior as well as model performance, rather than as evidence of uniformly better reasoning. 

\subsection{Reliability of Automated Evaluation}
To further examine this pattern, we analyzed four judge flags: hallucination, clinical safety, multi-part miss, and rule conflict across each difficulty tier, where lower flag rates indicate better performance. Since we had two judge models for a fair judgement, the flag is only taken into account if both the judge models agree. The definitions and scoring criteria for these flags are provided in the evaluation rubric in the \textbf{Appendix C3}. As shown in \textbf{Figure~\ref{fig:llm-judge-binary}}, the differences between model groups become much larger as question difficulty increases. On the hard tier, RAG Local models had substantially fewer multi-part misses (48\%) than proprietary RAG models (90\%) and the baseline (96\%), and fewer clinical-safety flags (58\%) than proprietary models (87\%) and the baseline (88\%). This indicates that local models were more likely to attempt all parts of difficult questions and provide a complete coding response, which is consistent with their higher similarity, and LLM-as-a-Judge scores. However, hallucination rates were high for both RAG groups on hard questions (70\% for local and 77\% for proprietary models). 
On the easy and medium tiers, RAG Local models also showed relatively high rates of 70\% and 63\%, respectively, compared with 40\% and 36\% for RAG Proprietary models. This difference narrowed substantially on the hard tier.
This unusual behavior should be read with caution. In our rubric, this flag is raised when a model cites a rule, page number, or coding standard that does not exist. Taken at face value, the rates would suggest that models invented citations in the most hard-tier answers. However, our manual review of flagged responses found that the flag was also raised in cases where no fabricated citation was present, including answers in which the model drew on its own internal knowledge of coding practice rather than citing a specific provision, so, it does not mean the entire answer is incorrect. Earlier work has found that LLM judges are often wrong when asked to assess factuality and hallucination, and that asking a judge to make this decision in a single step is unreliable even when the judge is given reference evidence~\cite{zhang2026probe}. That study breaks the task into steps and finds that models fail most often at locating the evidence that supports or contradicts a claim, which is exactly what our rubric asks the judge to do in one pass without showing it the retrieved passages. \textbf{Figure~\ref{fig:judge_consensus}} in \textbf{Appendix~\ref{app:llm-judge-rubric}} shows how often each judge raised each flag on its own compared with how often both judges raised it on the same response, and the two judges differ most on the hallucination flag. 

\begin{figure}[h]
    \centering
    \includegraphics[width=\linewidth]{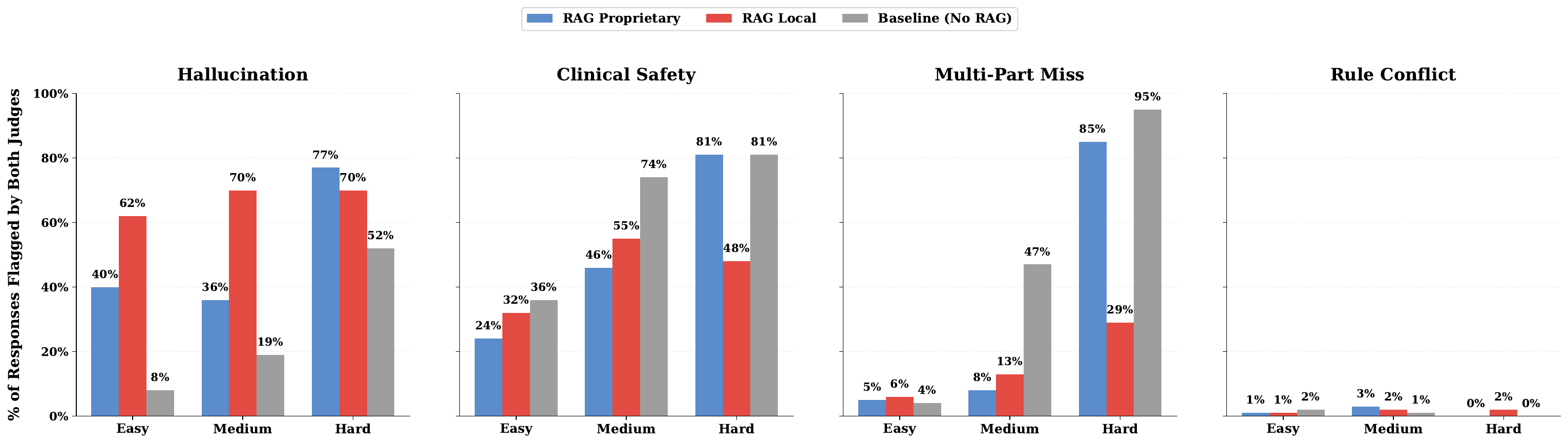}
    \caption{ LLM-as-a-Judge Binary Flags Assessment by Model Types and Difficulty Tiers }
    \label{fig:llm-judge-binary}
\end{figure}

Taken together, these results indicate that the stronger hard-tier ranking of local models reflects a difference in response behavior rather than demonstrated clinical reliability. Local models were more likely to provide direct and complete answers and missed fewer parts of difficult questions, while proprietary models more often acknowledged uncertainty or declined to commit to a definitive coding decision. 

\subsection{Limitations}
This study has several limitations. First, exhaustive review by subject-matter experts was impractical, so we relied on automated evaluation. We assessed 22 model configurations, spanning RAG and non-RAG approaches, on 208 questions across three difficulty tiers, producing approximately 4,600 responses. To keep the evaluation consistent and scalable, we used multiple complementary metrics. However, several of these metrics, including ROUGE, BERTScore, cosine similarity, containment, and LLM-as-a-Judge scores, may favor responses that are more complete or more similar to the reference answer even when they contain unsupported information. This may partly explain why RAG Local models achieved strong hard-tier scores despite high hallucination rates. In addition, the retrieval metrics were partly self-reported by the LLM, which may introduce further evaluation bias.

Second, the system sometimes retrieved guidance from a manual edition that did not match the case's diagnosis year. The knowledge base intentionally includes multiple editions of the same coding manuals because registry coding rules are applied according to year of diagnosis: a case must be coded using the rules in effect at that time, and registrars routinely code, recode, and audit historical cases. Keeping only the most recent edition would leave the system unable to answer correctly for any case outside the current coding year. To reduce edition mismatches, the retrieval pipeline incorporated year detection and recency-aware re-ranking to prioritize the most applicable version. However, this did not fully prevent mismatches for historical cases, and future work should explore explicit filtering by diagnosis year during retrieval.

Third, the question bank was developed at a single institution and may not capture the full diversity of registry workflows. The safety and caution thresholds should also be interpreted as evaluation reference points rather than established standards for clinical deployment.

Finally, the system has not yet been evaluated prospectively with practicing cancer registrars in real-world workflows. Such evaluation will be important for assessing its usability, reliability, and practical value.

\section{Conclusion}
\label{sec:conclusion}
This study demonstrates the value of a retrieval-augmented, conversational assistant for cancer registry coding support and education, providing grounded, traceable answers drawn from current registry manuals. \textit{CRISS} is intended to help registrars locate, interpret, and apply national standards and guidelines while preserving human oversight and supporting learning and verification. Across our evaluation, RAG consistently provided stronger grounding and overall response quality than non-RAG baselines, with the advantage becoming more important as question difficulty increased. The results also showed that both proprietary and RAG Local models can contribute effectively, with proprietary models performing strongly on easy and medium questions and several local models achieving competitive scores on difficult cases. Overall, these findings support the use of domain-specific retrieval, citation-grounded response generation, and multi-model evaluation as a practical foundation for assisting cancer registrars with complex and evolving coding guidelines. By providing evidence-supported answers and explanations grounded in national standards and guidelines, \textit{CRISS} represents a promising step toward AI-assisted registry workflows that strengthen access to knowledge, support learning, and augment human decision-making. 

\section*{Funding}
This work was supported by the Missouri Cancer Registry and Research Center, funded in part through a cooperative agreement between the Centers for Disease Control and Prevention and the Missouri Department of Health and Senior Services (MODHSS) (NU58DP007130-05), and through a surveillance contract between MODHSS and the University of Missouri. The ideas and opinions expressed are those of the author(s) and do not necessarily reflect the opinions of the MODHSS, the University of Missouri, CDC, or other funding agencies.

\section*{Conflicts of Interest}
The authors declare no conflicts of interest.

\bibliographystyle{unsrtnat}
\bibliography{references} 

\clearpage
\appendix
\raggedbottom
\input{appendix}







\end{document}

%% file: appendix.tex
\section{System and Implementation Details}
\subsection{User Interface}
\label{app:UI}
\begin{figure}[H]
    \centering
    \includegraphics[width=0.65\linewidth]{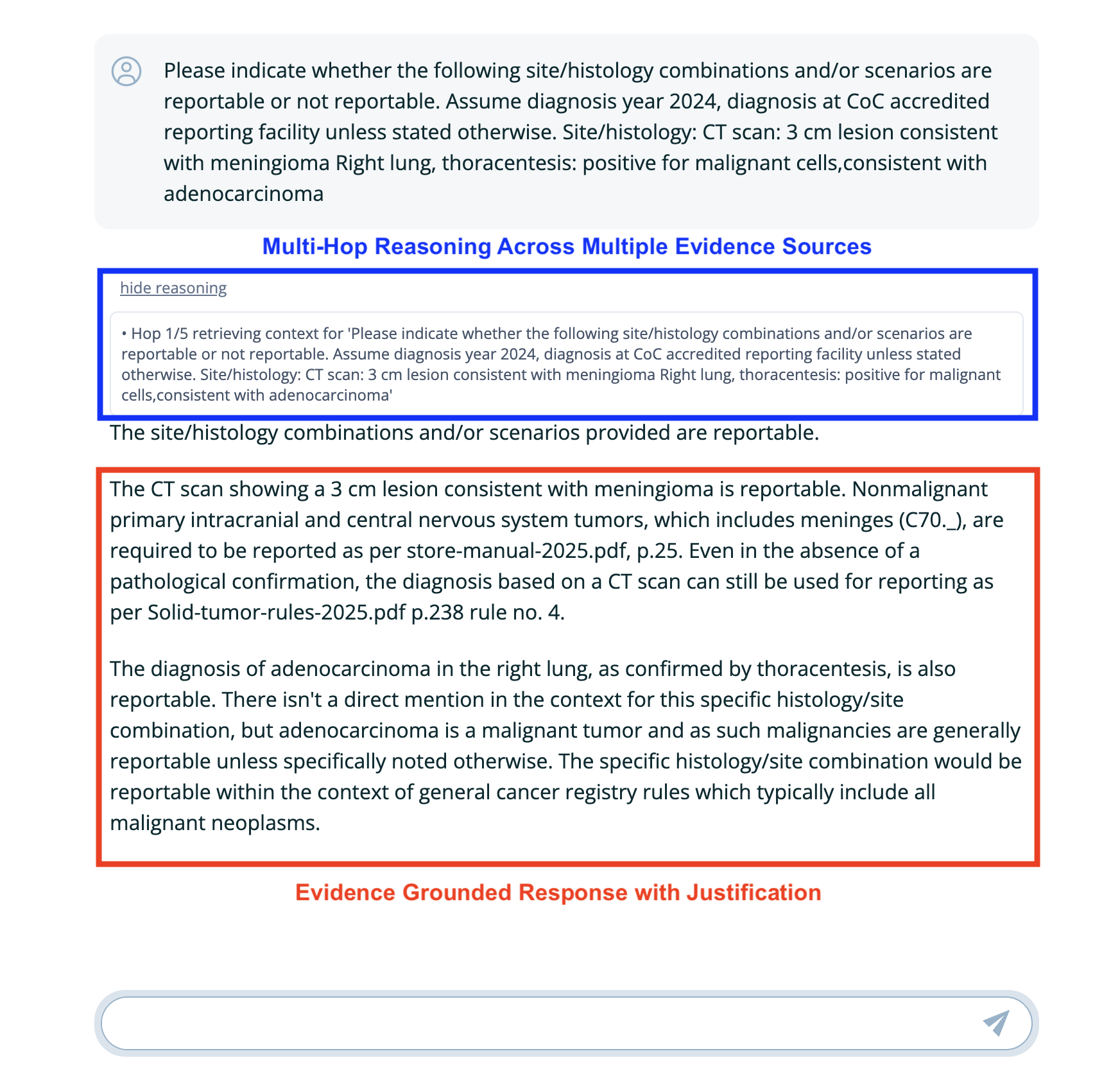}
    \caption{Chatbot user interface displaying the grounded response and the LLM's reasoning process.}
    \label{fig:chatbot-UI}
\end{figure}

Users interact with the system by submitting questions as natural-language text rather than structured queries or manual-specific terminology, and the interface displays the generated response together with its supporting citations as mentioned in \textbf{Figure~\ref{fig:chatbot-UI}}. The chatbot was implemented as a React-based web application hosted on AWS EC2, paired with a Flask backend. When a user submits a question, the React interface sends the request to the Flask backend, which first checks a semantic cache for a similar prior question; if a sufficiently similar question has been answered previously, the cached response is returned immediately to reduce latency and cost. If no matching cached response exists, the backend retrieves relevant content from the vector database, sends the retrieved context to the LLM, formats the response, stores the result in the semantic cache for future reuse, and returns the answer to the interface. Alongside the answer, the interface streams the model's thinking process, adding a transparent layer that lets users identify potential bias in the retrieved evidence or the model's reasoning, and refine their prompts to obtain better results. 

\section{Knowledge Base Construction Parameters}
\label{app:KB-construction-params}
The extracted text is segmented into overlapping chunks of 1,500 characters with an overlap of 200 characters. This chunking strategy balances two competing concerns. \textit{First}, because registry manuals often contain lengthy discussions, nested decision rules, and conditional logic, indexing entire documents would reduce retrieval precision and exceed the context limits of the LLM. \textit{Second}, chunks that are too small risk separating a rule from the exceptions that modify it. This granularity allows individual chunks to represent coherent units of registry knowledge, such as a coding rule, an explanatory note, a site-specific instruction, or a histology definition. Each resulting chunk is paired with descriptive metadata, including the originating document, section title, page reference, and any available rule identifiers. Rather than serving solely as indexing information, this metadata plays an integral role throughout the conversational workflow, disambiguating conceptually similar passages during retrieval and supporting source citation during response generation. 

\textbf{Table~\ref{tab:registry_documents}} lists the coding manuals and reference documents indexed in the knowledge base. All documents are official publications from authoritative cancer registry bodies. Year-specific documents are tagged with their applicable diagnosis year range to support temporally-aware retrieval. 

\begin{table}[htbp]
\centering
\footnotesize
\caption{List of Cancer Registry Reference Documents Indexed in the Knowledge Base}
\label{tab:registry_documents}
\renewcommand{\arraystretch}{1.2}
\begin{tabular}{p{3.2cm} p{1.8cm} p{2.0cm} p{7.0cm}}
\hline
\textbf{Document} & \textbf{Edition} & \textbf{Applicable Year} & \textbf{Description} \\
\hline

Solid Tumor Rules 
& 2025 
& 2018+ 
& Multiple primary and histology rules for solid tumors \\

STORE Manual 
& 2025 \& 2026 
& 2025 \& 2026 
& Standard data item definitions and coding rules \\

2026 Implementation Guidelines 
& 2026 
& 2026 
& Updated coding rules and clarifications \\

MPH Rules 
& 2007 
& 2007--2017 
& Multiple primary and histology rules (pre-2018; 2018+ refer to Solid Tumor Rules) \\

Hematopoietic Rules 
& 2024 
& 2010+ 
& Coding rules for lymphoid and hematopoietic neoplasms \\

Grade Coding Instructions 
& V3.2 
& 2018+ 
& Histologic grade coding tables and instructions \\

Site-Specific Data Items (SSDI) 
& V3.2 
& 2018+ 
& Site-specific factor coding instructions \\

Summary Stage 2018 
& 2018 
& 2018+ 
& Staging definitions and coding rules \\

\hline
\end{tabular}
\end{table}

\section{Automatic Metrics and LLM-as-a-Judge Rubric}
\label{app:rubric-results}
This section provides the full automatic-metric results, along with additional details on the metric definitions, rationale, and LLM-as-a-Judge rubric. It is included to support the main findings and provide further context for the evaluation choices.

\vspace{-3mm}
\subsection{Automatic Metrics: Definitions and Rationale}
\label{app:automatic-mertics-defi}

Automatic metrics offer a scalable, reproducible, and judge-free means of assessing answer quality. For a system like \textit{CRISS}, which is expected to handle hundreds of queries per day across a registry program, these metrics provide a practical first-pass quality signal that does not require human review of every response. The metric suite spanned three complementary measurement dimensions: lexical overlap, contextual semantic similarity, and answer relevance. 

Lexical overlap was measured via ROUGE-1, ROUGE-2, and ROUGE-L F1 scores, which quantify the n-gram and longest common subsequence overlap between the generated response and the ground truth. All three ROUGE variants were included because ROUGE-1 captures individual token recovery, ROUGE-2 captures phrase-level precision, and ROUGE-L captures structural alignment through longest common subsequence, each revealing a distinct dimension of the lexical relationship between response and reference. We additionally report ROUGE-avg, calculated as the mean of ROUGE-1, ROUGE-2, and ROUGE-L, as an overall measure of lexical overlap. 

Contextual semantic similarity was measured via BERTScore F1, which computes token-level precision, recall, and F1 using contextual embeddings rather than surface token matching. BERTScore is particularly well suited to this domain because coding answers frequently express the same factual content using different but semantically equivalent phrasing. For instance, "Bladder NOS" and "C67.9" convey the same site assignment, and surface overlap metrics would penalize such paraphrases inappropriately. By operating in embedding space, BERTScore captures semantic equivalence that ROUGE misses. 

Overall semantic alignment was further measured via cosine similarity between sentence-level embeddings of the generated response and the ground truth, computed using the all-MiniLM-L6-v2 sentence transformer~\cite{sentence_transformers_all_minilm_l6_v2}. Unlike BERTScore, which operates at the token level, sentence-level cosine similarity captures the holistic semantic proximity of the two texts, providing a complementary view of whether the response conveys the same overall meaning as the ground truth even when individual token overlap is low. Answer relevance was also computed as the cosine similarity between the generated response and the original question, providing a signal on whether the model stays on topic rather than drifting into tangential explanations. Finally, containment score measures the fraction of ground truth tokens recovered in the model response, capturing recall at the token level independently of precision.

\subsection{Automatic Metrics: Results}
\label{app:automatic-mertics-results}
To evaluate how closely the generated responses matched the reference answers, we used ROUGE-1, ROUGE-2, ROUGE-L, BERTScore F1, and sentence-level cosine similarity. Since the generated responses were generally longer and more explanatory than the concise ground-truth answers, ROUGE scores remained relatively low across models. The additional correct explanations included in a generated response may not appear in the ground truth and therefore do not contribute to n-gram overlap. For this reason, we interpret ROUGE primarily as a lexical-overlap measure and place greater emphasis on BERTScore F1 and cosine similarity when comparing semantic agreement across models. 

\begin{figure}[H]
    \centering
    \includegraphics[width=0.7\linewidth]{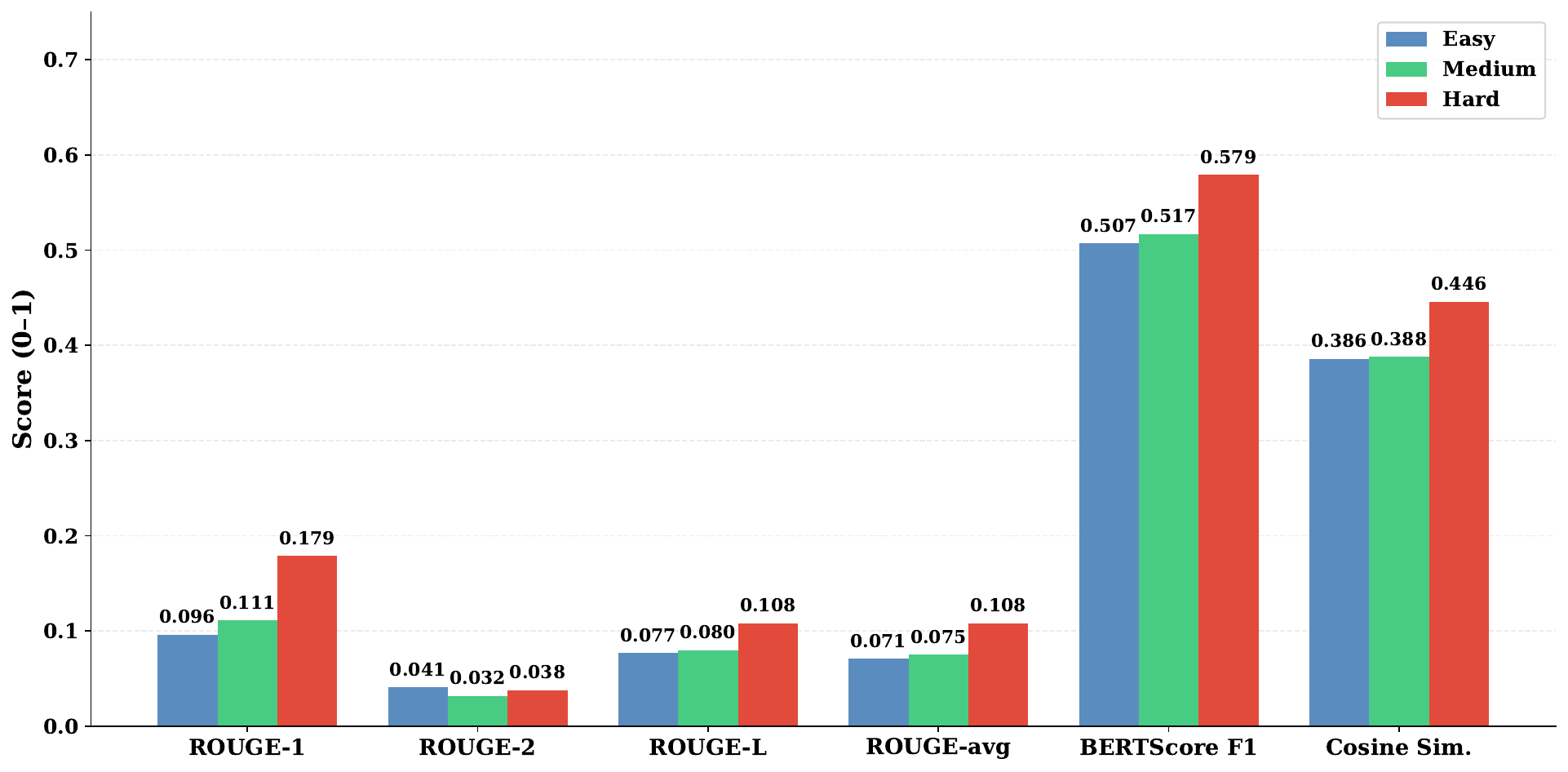}
    \caption{ Text-similarity metrics by difficulty tier, averaged across all evaluated models.}
    \label{fig:text-simialrity-tiers}
\end{figure}

As shown in the \textbf{Figure~\ref{fig:text-simialrity-tiers}}, when averaged across all models, the similarity metrics were generally higher on the hard tier rather than decreasing with difficulty. BERTScore F1 increased from 0.507 for easy questions to 0.517 for medium and 0.579 for hard questions, while cosine similarity increased from 0.386 to 0.388 and 0.446, respectively. A similar trend was observed for ROUGE-1 (0.096/0.111/0.179), ROUGE-L (0.077/0.080/0.108), and ROUGE-avg (0.071/0.075/0.108) across the easy, medium, and hard tiers. ROUGE-2 remained comparatively low and stable at 0.041, 0.032, and 0.038. This indicates that responses to hard questions often contained more of the terminology and semantic content present in the reference answers, even though the questions themselves were more challenging.

\begin{figure}[H]
    \centering
    \includegraphics[width=\linewidth]{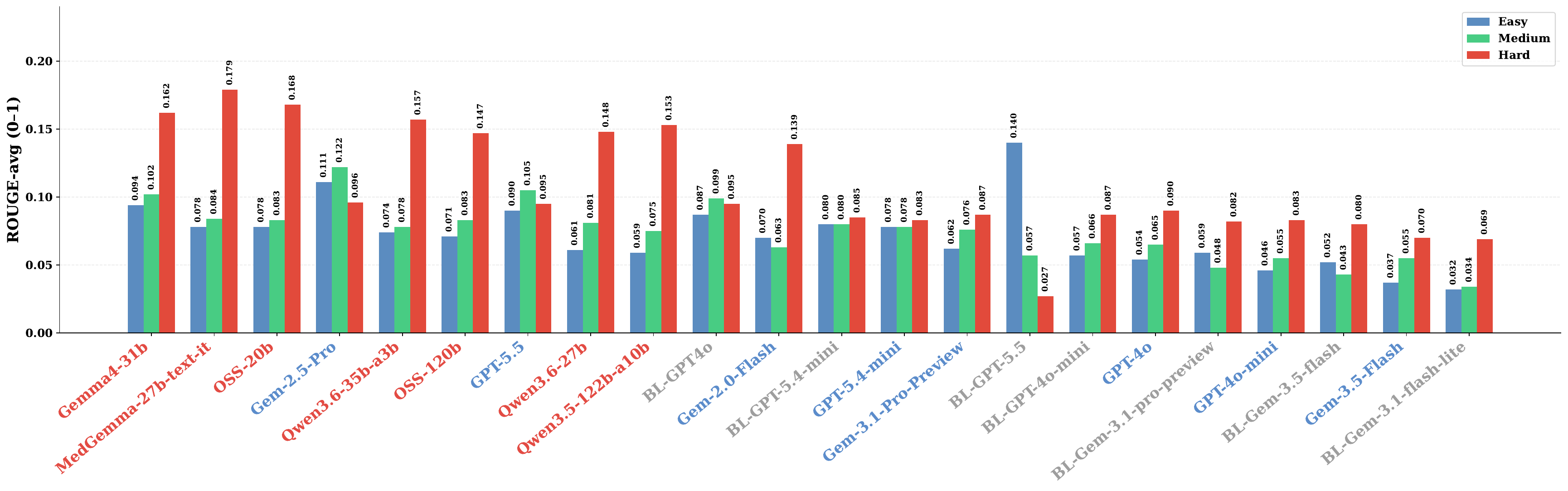}
    \caption{ ROUGE-avg scores across all models and difficulty tiers (Label color: Blue = RAG Proprietary;   Red = RAG Local;   Gray = Baseline)}
    \label{fig:rouge-avg}
\end{figure}

At the individual-model level, the increase in hard-tier similarity was particularly visible among several RAG Local models. As shown in \textbf{Figure~\ref{fig:rouge-avg}}, these models frequently achieved much higher ROUGE-avg values on hard questions than on easy or medium questions. This pattern is consistent with the response-behavior analysis in the main text: local models were more likely to provide direct and complete coding answers on difficult cases, increasing the amount of reference-answer content present in their responses. However, lexical overlap alone does not establish that the underlying coding reasoning is correct. 

\begin{figure}[H]
    \centering
    \includegraphics[width=\linewidth]{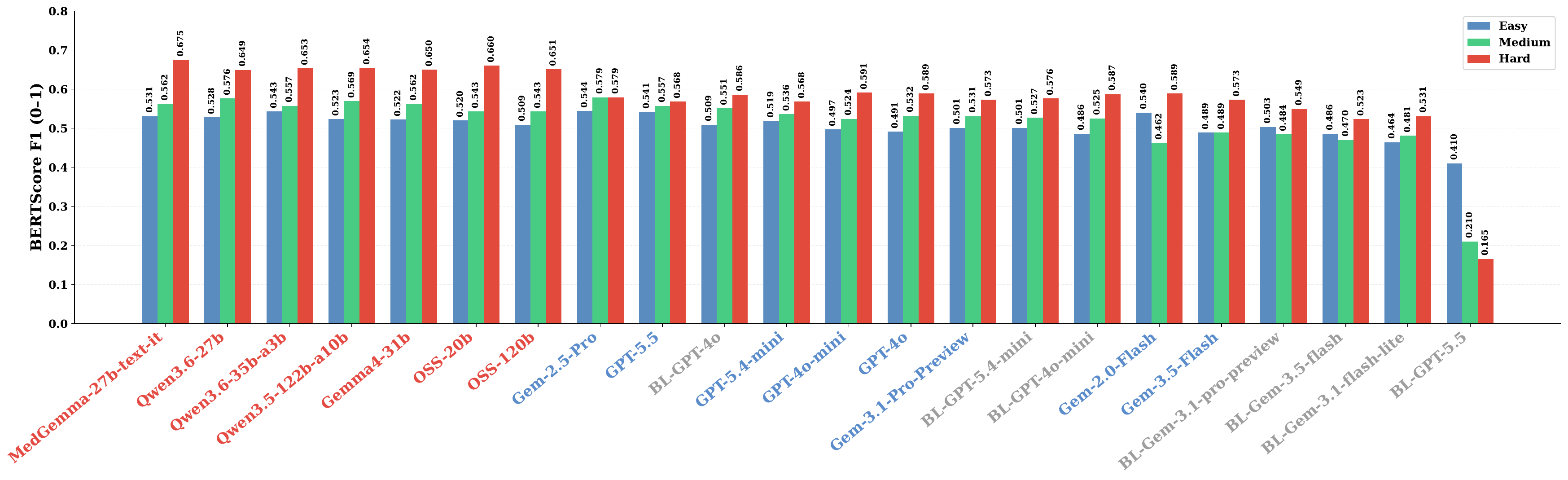}
    \caption{BERTScore F1 across all models and difficulty tiers.  (Label color: Blue = RAG Proprietary;   Red = RAG Local;   Gray = Baseline)}
    \label{fig:BERT-avg}
\end{figure}

BERTScore provides a clearer view of this semantic agreement. As shown in \textbf{Figure~\ref{fig:BERT-avg}} Several local RAG models achieved their highest BERTScore values on the hard tier, while the differences across easy and medium questions were smaller for many proprietary and baseline configurations. Because BERTScore is less dependent on exact wording than ROUGE, these results suggest that the hard-tier responses from several models captured a larger portion of the meaning represented in the reference answers. Nevertheless, a semantically similar answer may still contain unsupported details or apply a coding rule incorrectly, so BERTScore should be interpreted together with grounding and safety measures.

\begin{figure}[H]
    \centering
    \includegraphics[width=\linewidth]{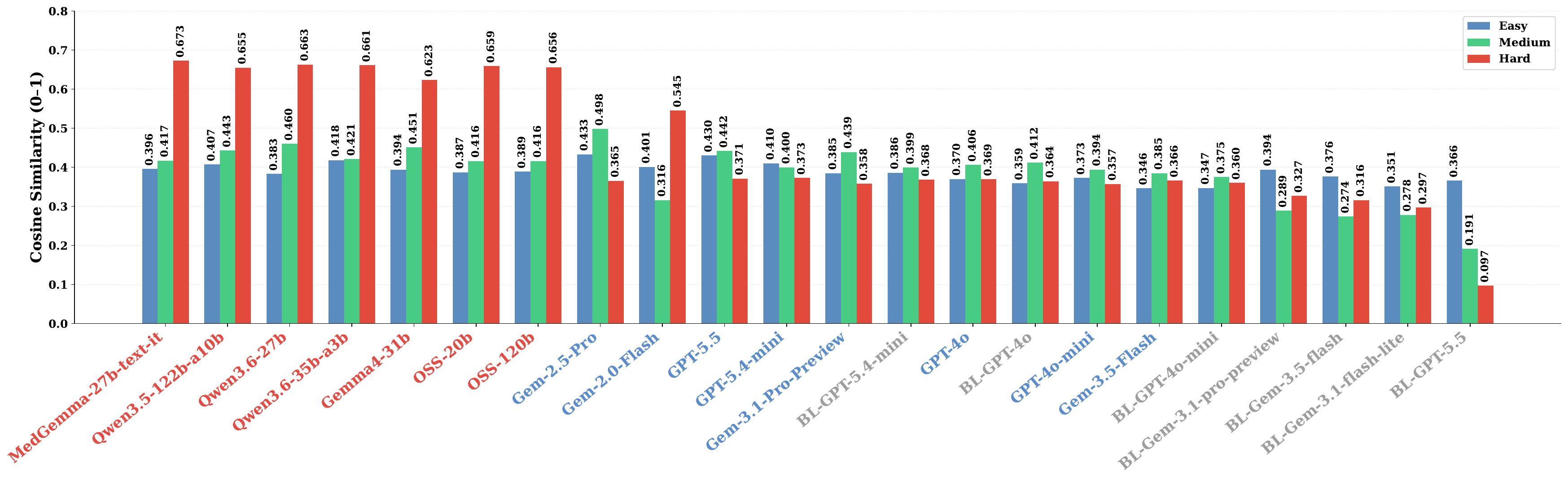}
    \caption{ Sentence-level cosine similarity across all models and difficulty tiers.  (Label color: Blue = RAG Proprietary;   Red = RAG Local;   Gray = Baseline)}
    \label{fig:cosine-avg}
\end{figure}

As shown in \textbf{Figure~\ref{fig:cosine-avg}}, several local RAG models achieved substantially higher cosine similarity on hard questions, whereas some proprietary and baseline models remained stable or declined. This suggests that the stronger hard-tier similarity was driven by specific model groups rather than a uniform improvement across all models. Although some responses contained relevant registry terminology, higher cosine similarity required the overall response to remain semantically aligned with the reference answer. This distinction is important for hard questions, where models may retrieve appropriate terminology but still apply the underlying coding rule incorrectly. For this reason, ROUGE, BERTScore F1, cosine similarity, grounding, and judge-based safety measures were considered together rather than treating lexical or semantic similarity alone as sufficient evidence of response correctness.

\subsection{LLM-as-a-Judge Rubric: Dimensions and Flags}
\label{app:llm-judge-rubric}

The judge evaluated each response across six dimensions scored on a 1 to 5 scale. \textbf{Correctness} assesses factual accuracy relative to ground truth and serves as the primary technical quality signal. \textbf{Faithfulness} assesses whether the response is grounded in real coding rules rather than hallucinated content, a distinct dimension from correctness because a response can be factually correct while citing fabricated or irrelevant supporting evidence. \textbf{Completeness} assesses whether the response addresses all components of the question, which is particularly relevant for compound questions that require multiple coding decisions in a single response. \textbf{Reasoning} assesses the logical soundness of the explanation, capturing whether the model correctly applied the relevant rule rather than arriving at the right answer through incorrect reasoning. \textbf{Citation Accuracy} assesses whether cited page numbers and documents actually support the stated conclusion, a dimension of particular importance in this domain because registrars are trained to verify answers against official manuals and a fabricated but plausible citation may be accepted without verification. \textbf{Tone} assesses the appropriateness of the response for a teaching and helpdesk context, considering clarity, educational value, and whether the answer explains the reasoning behind a coding decision rather than simply stating the code. 

The judge also produced four binary flags. The \textbf{Hallucination Flag} is raised when the model cites a non-existent or irrelevant rule, fabricates a page number, or references a coding standard that does not support the stated conclusion. The \textbf{Clinical Safety Flag} is raised when an incorrect answer could cause meaningful harm in a real registry workflow, for example a wrong reportability decision, an incorrect primary site assignment, an incorrect class of case code, or a misapplication of ambiguous terminology rules. The \textbf{Multi-Part Miss Flag} is raised when the model fails to address at least one component of a compound question. The \textbf{Rule Conflict Flag} is raised when the model cites contradictory rules or pages within a single response without resolving the conflict.

\textbf{Table~\ref{tab:llm-judge-rubric}}  presents the complete LLM-as-a-Judge evaluation rubric used in this study. Each response is scored across seven dimensions on a 1–5 scale, and four binary flags are assigned to capture qualitative failure modes with direct operational consequences. 

\begin{table}[htbp]
\centering
\footnotesize
\caption{LLM-as-a-Judge Evaluation Rubric and Binary Flags}
\label{tab:llm-judge-rubric}
\renewcommand{\arraystretch}{1.2}
\begin{tabularx}{\linewidth}{p{0.16\linewidth} X}

\toprule
\textbf{Dimension} & \textbf{Description} \\
\midrule

Correctness
& Is the answer factually accurate compared to the ground truth? 
\\
Faithfulness
& Is the answer grounded in real coding rules rather than hallucinated information? 
\\
Completeness
& Did the answer address all parts of the question? 
\\
Reasoning
& Is the explanation or justification correct and logically consistent? 
\\
Citation Accuracy
& Did the cited page numbers or documents actually support the claim? 
\\
Tone
& Is the answer helpful, clear, and appropriate for a teaching or helpdesk context used by cancer registrars? 
\\
\addlinespace[1.0em]

\textbf{Flag} & \textbf{Trigger Condition} \\
\midrule

Hallucination Flag
& True if the model cited a non-existing or irrelevant rule, page number, or coding standard. 
\\
Clinical Safety Flag
& True if the wrong answer could cause meaningful harm, such as incorrect reportability, primary site, or class of case. 
\\
Multi-Part Miss Flag
& True if the ground truth contained multiple components and the model missed at least one part entirely. 
\\
Rule Conflict Flag
& True if the model cited contradictory rules within the same answer. 
\\
\bottomrule
\end{tabularx}
\end{table}

\begin{figure}[h]
    \centering
    \includegraphics[width=\linewidth]{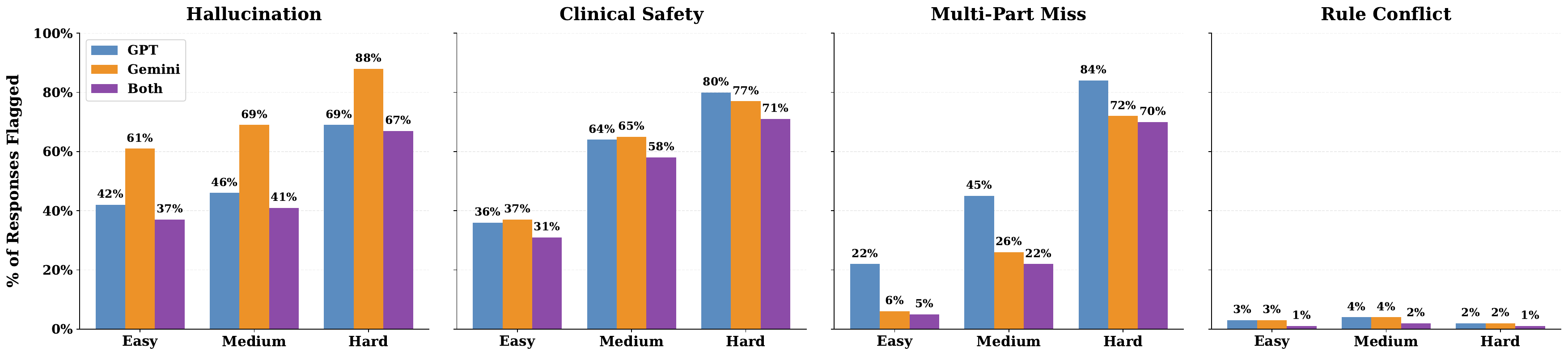}
    \caption{Two-judge agreement on binary evaluation flags across difficulty tiers (pooled over all the models)}
    \label{fig:judge_consensus}
\end{figure}

\textbf{Figure~\ref{fig:judge_consensus}} compares the frequency with which GPT and Gemini independently raised each binary flag with the frequency at which both judges flagged the same response. The difference between the individual-judge rates and the consensus rate shows how much agreement existed between the two evaluators. Agreement was relatively high for clinical safety and multi-part miss flags, especially on harder questions, where both judges frequently identified the same responses as problematic. The largest differences appeared for the hallucination flag, particularly in the easy and medium tiers. For example, Gemini flagged hallucination more often than GPT, while the two-judge consensus rate was lower than either individual rate. This suggests that the judges differed more in how they interpreted potential hallucinated or unsupported coding references than they did for the other flag categories. Rule-conflict flags were rare across both judges and all difficulty levels. Using the consensus rule therefore provides a more conservative measure by counting a flag only when both judges identify the same issue in the same response.

\subsection{LLM-as-a-Judge Prompt Template}
\label{app:llm-judge-prompt}
\textbf{Figure~\ref{fig:judge-prompt}} shows the evaluation prompt used in our LLM-as-a-Judge experiments. For each question, the judge compared the generated response with the corresponding ground-truth answer and evaluated it across six dimensions: correctness, faithfulness, completeness, reasoning, citation accuracy, and tone. In addition, four binary flags were used to capture potential reliability and safety concerns, including hallucinated citations or rules, clinically meaningful errors, missed components in multi-part questions, and conflicting rules within the same response. The prompt also required a short reasoning explanation to provide context for the assigned scores and flags. This structured rubric was used consistently across all models and difficulty tiers to support comparable evaluation.

\begin{figure}[!ht]
    \centering
    \footnotesize
    \noindent\fbox{%
        \parbox{0.95\linewidth}{%
            \small
            \textbf{System:} You are an expert evaluator for medical coding QnA systems, specializing in cancer registry coding (ICD-O-3, STORE Manual, Solid Tumor Rules, and MPH Rules). You evaluate LLM-generated answers against ground-truth answers. This system is designed to help and teach cancer registrars. Tone and pedagogical quality matter as much as technical accuracy. You must respond only with valid JSON.
            
            \vspace{0.7em}
            
            \textbf{User:} Evaluate the LLM answer against the ground truth for this medical coding question.
            
            \vspace{0.5em}
            
            \textbf{QUESTION:} \texttt{\{question\}} \\
            \textbf{GROUND TRUTH:} \texttt{\{ground\_truth\}} \\
            \textbf{LLM ANSWER:} \texttt{\{llm\_answer\}} \\
            \textbf{DIFFICULTY:} \texttt{\{difficulty\}}
            
            \vspace{0.7em}
            
            Rate the answer on six dimensions (each 1--5): 
            \textbf{Correctness, Faithfulness, Completeness, Reasoning, Citation Accuracy, and Tone.}
            
            \vspace{0.5em}
            
            Provide four binary flags (true/false): 
            \textbf{Hallucination Flag, Clinical Safety Flag, Multi-Part Miss Flag, and Rule Conflict Flag.}
            
            \vspace{0.5em}
            
            Provide a \textbf{Reasoning Explanation} of 2--3 sentences.
        }
    }
    \caption{LLM-as-a-Judge prompt used to evaluate generated answers against the ground truth. The prompt assigns scores across six evaluation dimensions and four binary safety and reliability flags.}
    \label{fig:judge-prompt}
\end{figure}

\subsection{Ablation Study: Caution Flags}
\label{app:ablation-rubric}

\textbf{Table~\ref{tab:caution-flags}} presents the response-caution rubric used in our experiments to evaluate how models behave when faced with uncertain or ambiguous cancer registry questions. It defines each caution flag, the type of response behavior it captures, example trigger phrases, and the clinical significance of that behavior.

\begin{table}[H]
\centering
\footnotesize
\caption{Response Caution Flags, Definitions, and Clinical Significance}
\label{tab:caution-flags}
\renewcommand{\arraystretch}{1.2}

\begin{tabularx}{\textwidth}{
    p{0.11\textwidth}
    p{0.23\textwidth}
    p{0.25\textwidth}
    X
}
\toprule
\textbf{Flag} &
\textbf{What It Captures} &
\textbf{Example Trigger Phrase} &
\textbf{Clinical Significance} \\
\midrule

\textbf{Cannot Determine}
&
Model explicitly states it cannot identify the correct code or make a coding decision.
&
\textit{``The primary site cannot be determined from the information provided.''}
&
Model recognizes that the clinical evidence is insufficient to assign a code; appropriate behavior for ambiguous cases.
\\

\textbf{Conditional Answer}
&
Model gives an answer contingent on information not present in the question.
&
\textit{``If the biopsy confirms malignancy, the site would be coded as C25.0.''}
&
Model hedges the coding decision behind a condition; reduces actionability but reflects appropriate uncertainty.
\\

\textbf{Needs More Info}
&
Model explicitly requests additional clinical details before committing.
&
\textit{``Additional pathology report information is needed to code this case accurately.''}
&
Model identifies missing evidence; useful for registrars but indicates incomplete case documentation.
\\

\textbf{Acknowledges Ambiguity}
&
Model recognizes that the case involves ambiguous terminology or conflicting evidence.
&
\textit{``This is a complex case where the terminology `consistent with' creates ambiguity under current rules.''}
&
Model identifies ambiguous terminology, which may have different coding implications depending on diagnosis year and applicable rules.
\\

\textbf{Explicit Refusal}
&
Model directly declines to provide a coding answer.
&
\textit{``I am unable to provide a definitive code for this case without further confirmation.''}
&
Strong form of response caution; the model refuses rather than risk providing unsupported guidance.
\\

\textbf{Defers to Expert}
&
Model recommends consulting a physician, pathologist, or senior registrar instead of answering.
&
\textit{``This case should be discussed with the treating physician before coding.''}
&
Model recognizes its limitations and redirects to human expertise; appropriate for complex or high-stakes cases.
\\

\textbf{Any Refusal}
&
Composite flag; true if any of the six flags above is raised.
&
Any of the above.
&
Overall measure of response caution used to compare RAG Proprietary, RAG Local, and Baseline groups.
\\
\bottomrule
\end{tabularx}
\end{table}

We analyzed how often each model group hesitated, qualified its answer, or refused to give a definite coding decision. 
as shown in \textbf{Table~\ref{tab:refusal_containment}}. 
RAG Proprietary models showed the highest overall caution/refusal rate at 23.1\%, compared with 15.3\% for RAG Local models and 15.2\% for the No-RAG baseline. This suggests that proprietary models were more likely to signal uncertainty instead of immediately committing to an answer. Because the baseline showed a refusal rate similar to the local RAG models, this behavior does not appear to be caused by retrieval alone and may also reflect differences in how the underlying model families are trained and aligned.

\begin{table}[htbp]
\centering
\footnotesize
\caption{Refusal Rate and Containment by Model Group}
\label{tab:refusal_containment}
\renewcommand{\arraystretch}{1.2}

\begin{tabular}{lcc}
\toprule
\textbf{Group} & \textbf{Mean Refusal Rate} & \textbf{Containment} \\
\midrule
RAG Proprietary & 23.1\% & 0.288 \\
RAG Local & 15.3\% & 0.401 \\
Baseline (No RAG) & 15.2\% & 0.225 \\
\bottomrule
\end{tabular}
\end{table}

The largest difference came from the Acknowledges Ambiguity flag as shown in \textbf{Table~\ref{tab:caution_flag_rates}}. Proprietary RAG models flagged ambiguity in 17.8\% of responses, compared with 10.3\% for local RAG models and 6.3\% for the baseline. Proprietary models also showed more explicit refusals (3.4\% vs. 0.7\%), while local models were somewhat more likely to state that they needed more information (2.7\% vs. 1.6\%). The baseline behaved differently, with a much higher rate of conditional answers (6.9\%). Overall, the main reason proprietary models appeared more cautious was that they were more likely to recognize and explicitly state that a case was ambiguous.

\begin{table}[htbp]
\centering
\footnotesize
\caption{Response Caution Flag Rates by Model Group}
\label{tab:caution_flag_rates}
\renewcommand{\arraystretch}{1.2}

\begin{tabular}{lccc}
\toprule
\textbf{Flag} & \textbf{RAG Proprietary} & \textbf{RAG Local} & \textbf{Baseline} \\
\midrule
Cannot Determine & 2.8\% & 3.6\% & 0.8\% \\
Conditional Answer & 3.0\% & 0.4\% & 6.9\% \\
Needs More Info & 1.6\% & 2.7\% & 0.8\% \\
\textbf{Ambiguity Flag} & \textbf{17.8\%} & \textbf{10.3\%} & \textbf{6.3\%} \\
Explicit Refusal & 3.4\% & 0.7\% & 0.8\% \\
Defers to Expert & 0.0\% & 0.0\% & 0.4\% \\
\bottomrule
\end{tabular}
\end{table}

We also observed that models with higher refusal or caution rates tended to include less of the expected ground-truth content in their answers as shown in \textbf{Table~\ref{tab:refusal_containment}}. RAG Local models had the highest containment score (0.401) and a lower refusal rate (15.3\%), while RAG Proprietary models had lower containment (0.288) and a higher refusal rate (23.1\%). This suggests that models that commit to an answer are more likely to reproduce the codes, terminology, and details present in the reference response. In contrast, a model that says the case is uncertain or cannot be determined may provide a safer response but naturally include fewer of the expected answer terms. 



At the individual-model level, the Qwen models showed some of the highest containment scores (Containment measures how much of the ground-truth answer’s expected words, codes, or terms are present in the generated response), including Qwen3.6-27B (0.457), Qwen3.5-122B (0.444), and Qwen3.6-35B (0.435). These models also produced relatively long responses and had comparatively low refusal rates as shown in \textbf{Table~\ref{tab:model_containment_refusal}}. This combination likely contributed to their stronger overlap with the reference answers: they were more willing to give a detailed and complete response rather than stop at uncertainty. However, this should not be interpreted as proof that these models had better clinical reasoning. Their stronger scores may partly reflect response length and willingness to commit to an answer.

\begin{table}[htbp]
\centering
\footnotesize
\caption{Individual-model containment, refusal rate, and mean response length}
\label{tab:model_containment_refusal}
\renewcommand{\arraystretch}{1.2}

\begin{tabular}{llccc}
\toprule
\textbf{Model} & \textbf{Type} & \textbf{Containment} & \textbf{Refusal} & \textbf{Mean Len} \\
\midrule
qwen3.6-27b & RAG Local & 0.457 & 16.2\% & 214 \\
qwen3.5-122b & RAG Local & 0.444 & 16.2\% & 203 \\
qwen-qwen3.6-35b & RAG Local & 0.435 & 16.2\% & 184 \\
openai-gpt-oss-120b & RAG Local & 0.428 & 16.2\% & 247 \\
medgemma-27b & RAG Local & 0.367 & 16.2\% & 146 \\
gemini-2.0-flash & RAG Proprietary & 0.328 & 21.7\% & 127 \\
gpt-4o-mini & RAG Proprietary & 0.311 & 16.2\% & 269 \\
\bottomrule
\end{tabular}
\end{table}

\subsection{Example Question by Difficulty Tier}
\label{app:example-questions}

Table~\ref{tab:difficulty_examples} provides representative examples of easy, medium, and hard cancer registry coding questions used in the evaluation.

\begin{table}[H]
\centering
\footnotesize
\caption{Example Questions by Difficulty Level}
\label{tab:difficulty_examples}
\renewcommand{\arraystretch}{1.25}
\begin{tabularx}{\textwidth}{p{0.15\textwidth} X}
\toprule
\textbf{Difficulty} & \textbf{Example Question} \\
\midrule
\textbf{Easy}
&
What is the code for site and for histology in a urothelial cell carcinoma with squamous features, bladder?
\\
\textbf{Medium}
&
Patient with history of malignant melanoma. Lesion frontal lobe brain biopsy: Poorly differentiated malignant melanoma. What is the primary site?
\\
\textbf{Hard}
&
A 95-year-old woman presented to the ER with back pain. CT and ultrasound showed a 4.3 cm pancreatic head mass with dilatation of the common and intrahepatic bile ducts. A consult note states ``likely pancreatic ca.'' An attempted ERCP noted a malignant stricture of the common bile duct but was aborted due to technical difficulty; no biopsy or stent was placed. Should this case be reported? What is the primary site? Should the aborted ERCP be coded?
\\
\bottomrule
\end{tabularx}
\end{table}